\documentclass[letterpaper]{article} 
\usepackage{pir}  
\usepackage{times}  
\usepackage{helvet}  
\usepackage{courier}  
\usepackage[hyphens]{url}  
\usepackage{graphicx} 
\usepackage{natbib}  
\usepackage{caption} 
\usepackage{algorithm}
\usepackage{algorithmic}
\usepackage{amsmath}
\usepackage{amsfonts}
\usepackage{booktabs}
\usepackage{tabularx}
\usepackage{multirow}
\usepackage{xcolor}
\usepackage{newfloat}
\usepackage{listings}
\DeclareCaptionStyle{ruled}{labelfont=normalfont,labelsep=colon,strut=off} 
\floatstyle{ruled}
\newfloat{listing}{tb}{lst}{}
\floatname{listing}{Listing}
\nocopyright

\title{Multi-modal Policies with Physics-informed Representations in Complex Fluid Environments}
\author{
    Haodong Feng\textsuperscript{\rm 1,2}, 
    Peiyan Hu\textsuperscript{\rm 3},
    Yue Wang\textsuperscript{\rm 4} \equalcontrib,
    Dixia Fan\textsuperscript{\rm 1} \equalcontrib \\
}
\affiliations{
    \textsuperscript{\rm 1} Westlake University, 
    \textsuperscript{\rm 2} Zhejiang University, \\
    \textsuperscript{\rm 3} Academy of Mathematics and Systems Science, Chinese Academy of Sciences,   \\
    \textsuperscript{\rm 4} Zhongguancun Academy  \\

    \{fenghaodong, hupeiyan, fandixia\}@westlake.edu.cn,  yuewang@bjzgca.edu.cn
}

\usepackage{bibentry}

\begin{document}

\maketitle

\begin{abstract}
Control in fluid environments is an important research area with numerous applications across various domains, including underwater robotics, aerospace engineering, and biomedical systems. However, in practice, control methods often face challenges due to sparse or missing observations, stemming from sensor limitations and faults. These issues result in observations that are not only sparse but also inconsistent in their number and modalities (e.g., velocity and pressure sensors).
In this work, we propose a \textbf{\underline{P}}hysics-\textbf{\underline{I}}nformed \textbf{\underline{R}}epresentation (PIR) algorithm for multi-modal policies of control to leverage the sparse and random observations in complex fluid environments.
PIR integrates sparse observational data with the Partial Differential Equation (PDE) information to distill a unified representation of fluid systems.
The main idea is that PDE solutions are determined by three elements: the equation, initial conditions, and boundary conditions. 
Given the equation, we only need to learn the representation of the initial and boundary conditions, which define a trajectory of a specific fluid system.
Specifically, it leverages PDE loss to fit the neural network and data loss calculated on the observations with random quantities and multi-modalities to propagate the information with initial and boundary conditions into the representations. The representations are the learnable parameters or the output of the encoder. In the experiments, the PIR illustrates the superior consistency with the features of the ground truth compared with baselines, even when there are missing modalities. Furthermore, PIR combined with Reinforcement Learning has been successfully applied in control tasks where the robot leverages the learned state by PIR faster and more accurately, passing through the complex vortex street from a random starting location to reach a random target. 
\end{abstract}


\section{Introduction}
The control task represents a critical frontier in engineering research, especially in the complex fluid environment, due to the challenges posed by its strong nonlinear, unstable, and partially observed features. The importance of fluid environment control is underscored by its wide-ranging applications across numerous disciplines, from fluid dynamics \citep{verma2018efficient} to plasma physics \citep{degrave2022magnetic} and particle dynamics \citep{reyes2023magnetic}. This field needs to develop control policies that can navigate the intricate dynamics of fluid systems. 

However, in practical applications, the control in complex fluid environments faces several significant challenges \citep{manohar2022sparse, shang2011sensor, loiseau2018sparse}, which impose important requirements on the control algorithm: 
(1) \textbf{Sparse observations}: 
The algorithm must be specifically designed to work with sparse observations, which is a common nature in practical fluid environment monitoring. In most cases, comprehensive data coverage across the entire domain is not feasible due to physical, economic, or technical constraints. The control algorithm must demonstrate the capability to make decisions based on limited information distributed across the domain of interest. 
(2) \textbf{Irregular and arbitrary number of sensors}: 
The control algorithm must be capable of handling inputs from an irregular and arbitrary number of sensors distributed across the fluid domain. 
This flexibility is crucial as the number and spatial placement of sensors generally vary significantly across different testing environments and operational scenarios. 
(3) \textbf{Random faults of different sensor types}: 
The algorithm must maintain robust functionality when sensors fail or data becomes unavailable, effectively handling partial multi-modal information (velocity and pressure sensors). This requirement is essential because sensors are prone to damage in complex environments due to physical wear and environmental interference.

\begin{figure*}[!htb]
\centering
\centerline{\includegraphics[width=0.92\textwidth]{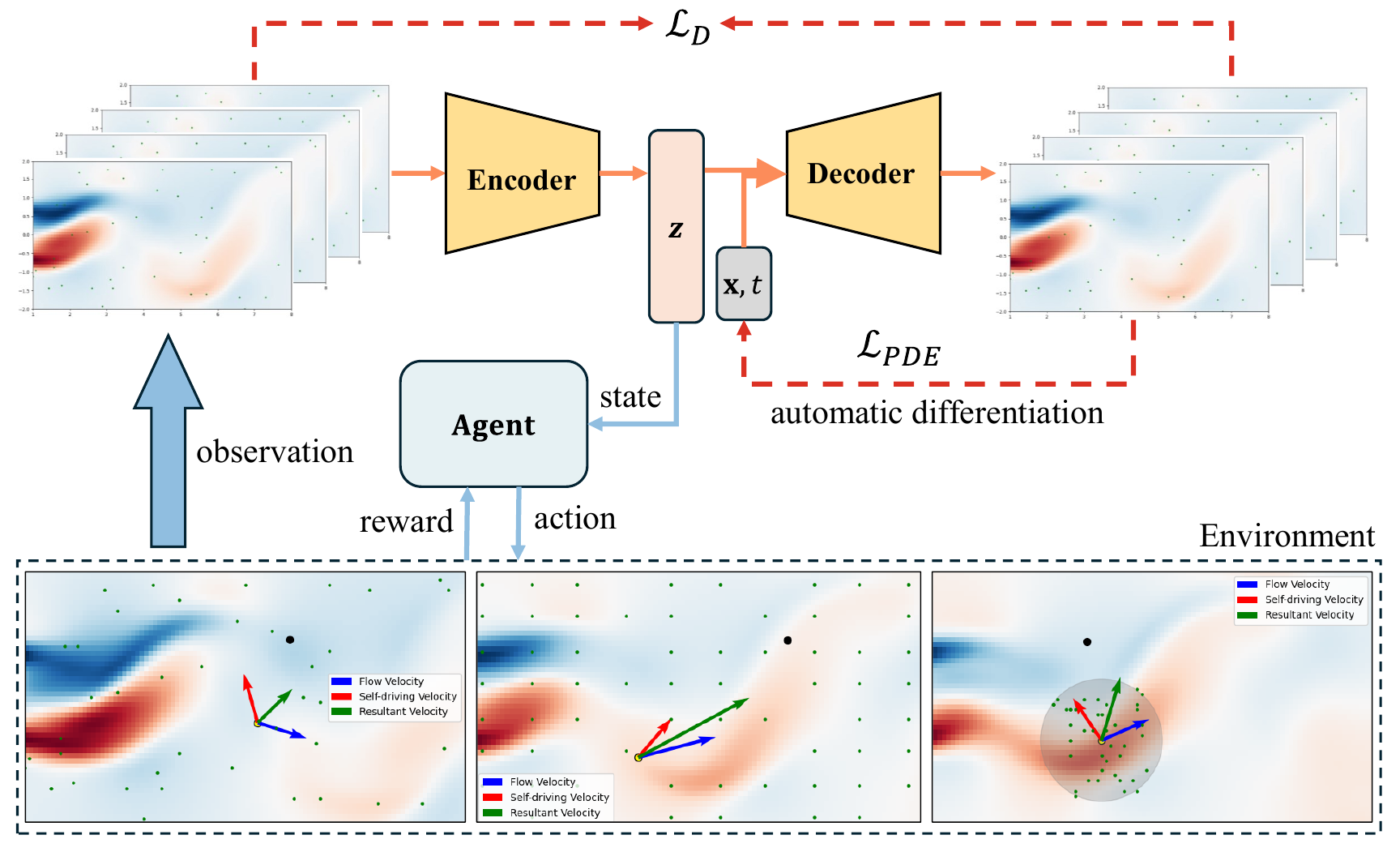}}
\caption{\textbf{The framework of PIR}. The framework consists of two components: an encoder and a physics-informed decoder. The red link denotes the calculation of data loss $\mathcal{L}_D$ and PDE loss $\mathcal{L}_{PDE}$. $\mathcal{L}_D$ is calculated using the input of the encoder and the output of the decoder so that the decoder can reconstruct the observation by using the representation $z$. $\mathcal{L}_{PDE}$ is calculated by the automatic differentiation of the neural network. The learned representation $z$ is input into the RL agent as state. The agent generates the actions that are executed in the environment, and then multi-modal sparse observations are obtained from the environment. 
The observations enter the encoder to generate a new representation.}
\label{fig.1}
\end{figure*}

To address these challenges, we propose a multi-modal control policy with \textbf{\underline{P}}hysics-\textbf{\underline{I}}nformed \textbf{\underline{R}}epresentations (PIR) that combines data-driven learning with the physical information of fluid dynamics. Rather than only relying on sensor observations, PIR integrates the fundamental physical laws that govern fluid behavior. Thus, physical information is used to implicitly establish inherent correlations between different time, space, and observation. PIR has three advantages. \textbf{First}, to manage the challenge caused by sparse observations, we leverage the PDE information in addition to the observed data, 
by applying a PDE loss to mitigate data sparsity limitations. 
\textbf{Second}, to handle the irregular and arbitrary number of sensors, we map observation data into a unified latent space representation, thereby standardizing the algorithm's inputs and accommodating varying sensor configurations across different operational environments. 
\textbf{Third}, to address random fault of different sensor types scenarios, we exploit the inherent relationships between different modalities in the PDEs to represent missing data. For instance, in the Navier-Stokes Equation (NSE), either pressure or velocity fields theoretically enables the inference of the other through the governing physical relationships. This modality-compensation approach ensures robust operation even when specific sensor types fail or become unavailable.

Specially, PIR leverages a decoder architecture inspired by Physics-informed Neural Networks (PINNs), which integrates the PDE formula directly into the learning process.
The core idea is that there are three elements that determine a PDE solution: the equation, initial conditions, and boundary conditions. When the equation is given, we only need to characterize the initial and boundary conditions to represent a definite fluid trajectory.
Therefore, PIR learns the initial and boundary conditions that are known in PINNs but unknown in this work by defining learnable representations as inputs to the decoder. Along with the learnable representations, the temporal information $t$ and spatial coordinates $x$ and $y$ are also inputs (shown in Fig. \ref{fig.1}). The outputs are the corresponding fluid fields, such as velocity components $u$ and $v$, and pressure $p$. By employing backpropagation, it calculates the PDE loss, which serves as a feedback mechanism for optimizing both the neural network parameters and the learnable representations. 
The optimization process ensures that the model not only fits the observed data but also adheres to the governing PDE.
Moreover, to make the algorithm more efficient, we incorporate an encoder to generate the representations. This modification significantly accelerates inference time during the control process, making the algorithms available to real-time feedback control. In this work, we combine the PIR and the Reinforcement Learning (RL) algorithm to achieve multi-modal control tasks in complex fluid environments. The proposed method is applied to efficiently navigate the robot in complex fluid vortex streets. The results indicate that our method outperforms existing baseline methods with an improvement from 20.0\% to 214.8\%. 

The contributions of our work are summarized as follows: (1) We introduce a novel physics-informed representation algorithm, PIR, which leverages both physical information and sparse observation to effectively learn the representation of trajectories within a fluid environment. (2) We combine the proposed PIR algorithm with RL, demonstrating its superior performance compared to baseline methods. PIR enhances the agent's ability to navigate and make decisions with sparse observation, an irregular and arbitrary number of sensors, and random faults of different sensor types. 


\section{Related Works} \label{related work}
\subsection{Physics-informed Machine Learning}
In recent years, many popular physics-informed machine learning methods have emerged. These methods embed known PDE information into the models, enabling them to solve the system of equations or achieve better generalization, particularly when data is scarce. PINNs \citep{raissi2017physics} train the neural networks with the PDE loss, which forces the output to minimize the residual of PDE. This work has led to applications across diverse domains, including fluid dynamics, solid mechanics, and mesh adaptation \citep{carleo2019machine, yang2020physics, hu2024better}. Building on PINNs, many have proposed various improved versions. For example, PINN-LS \citep{mowlavi2021optimal} improves training by treating the objective function as a regularization term with adaptive weights, CPINN \citep{zeng2022competitive} trains a discriminator to identify and correct errors, which enhances model accuracy, and Hodd-PINN exemplifies the integration of high-order numerical schemes into PINNs \citep{you2023high}, among others.

\subsection{Deep Learning-based Control in Fluid Environments}
Deep learning-based control methods can be categorized into three main types. Supervised learning methods train a surrogate model for forward prediction and then optimize control using gradients derived from backpropagation. For example, a hierarchical predictor-corrector approach was introduced for controlling complex fluid systems \citep{holl2020learning}, and later, a two-stage method was developed to learn the solution operator and apply gradient descent to determine optimal control \citep{hwang2022solving}. Reinforcement learning (RL) methods are also widely applied, with various algorithms such as Deep Q Learning and Proximal Policy Optimization (PPO) being utilized to control in fluid environments \citep{viquerat2022review, garnier2021review, gunnarson2021learning,zhang2025efficient,feng2024efficient}, as demonstrated by the open-source DRLinFluids platform \citep{0103113}. However, RL methods typically focus on maximizing rewards without considering physics-based information. Finally, PDE-constrained methods, such as PINN-based approaches, offer control through solving PDE without data \citep{MOWLAVI2023111731, barrystraume2022physicsinformed}. Recently, new methods have emerged, and they utilized advanced models for control, such as diffusion models, to simultaneously generate control sequences and state trajectories for control tasks, achieving promising results \citep{wei2024generative, wei2024closed, hu2024wavelet, hu2025uncertain}.

\begin{algorithm}[t]\small
   \caption{SAC with PIR}
   \label{alg.1}
    \begin{algorithmic}
   \STATE {\bfseries Input:} Dataset $\mathcal{D}$, Parameters $\theta$, $\alpha$, $\phi$, $\psi$, Number of episodes for PIR $N$ and SAC $M$.
   \STATE \textbf{Initialize} $\theta$ and $\alpha$ of encoder and decoder, policy $\pi_\phi$, critic $Q_\psi$, target critic $Q_{\psi'}$, replay buffer $\mathcal{B}$.
   \FOR{$episode = 1$ to $N$}  
   \STATE Calculate $\mathcal{L}_D$ and $\mathcal{L}_{PDE}$.
   \STATE Update $\theta$ and $\alpha$ using $\mathcal{L}_D$.
   \STATE Update $\theta$ using $\mathcal{L}_{PDE}$.
   \ENDFOR
   \STATE Set encoder parameters $\alpha$ to non-trainable.
   \FOR{$episode = 1$ to $M$}
    \STATE Reset environment, get initial state $s_0$.
    \STATE Encode state: $z_0 = E_\alpha(s_0)$.
    \FOR{$t = 0$ to $T-1$}
    \STATE Sample action: $a_t \sim \pi_\phi(\cdot|z_t)$.
    \STATE Execute action, get reward $r_t$ and next state $s_{t+1}$.
    \STATE Encode next state: $z_{t+1} = E_\alpha(s_{t+1})$.
    \STATE Store transition $(z_t, a_t, r_t, z_{t+1})$ in $\mathcal{B}$.
    \STATE Sample batch from $\mathcal{B}$ and update $\phi, \psi$.
   \ENDFOR
   \ENDFOR
\end{algorithmic}
\normalsize
\end{algorithm}

\section{Problem Setup} \label{background}
We consider the following widely used fluid systems:

\begin{equation}
\begin{aligned}
& \frac{\partial \mathbf{u}}{\partial t}+\mathcal{F}\left(\mathbf{u}, \nabla \mathbf{u}, \nabla^2 \mathbf{u}\right)  =0, \\
& \left.\mathcal{B}(\mathbf{u}, \nabla \mathbf{u})\right|_{\mathbf{x} \in \partial \Omega} =0, \\
& \left.\mathbf{u}\right|_{t=0} =\mathbf{u}_0,
\end{aligned}
\label{eqn.1}
\end{equation}
where $\mathbf{u}(t, \mathbf{x}):[0, T] \times \Omega \mapsto \mathbb{R}^{d_{\mathrm{u}}}$ is the trajectory of the system composed of states $\{\mathbf{u}(t, \cdot), t \in[0, T]\}$ defined on time range $[0, T] \subset \mathbb{R}$ and spatial domain $\Omega \subset \mathbb{R}^D$. $\mathcal{F}$ is a nonlinear operator that characterizes the dynamics of the PDE system, like the NSE. $\left.\mathcal{B}(\mathbf{u}, \nabla \mathbf{u})\right|_{\mathbf{x} \in \partial \Omega}$ is the boundary condition where $\mathcal{B}$ is a linear operator operating on the boundary $\partial \Omega$ of domain $\Omega$. The initial condition is specified by $\mathbf{u}_0(\mathrm{x})$. In such a system, we observe a finite training set of trajectories $\mathcal{D}$ given different ICs $u_0$, using the irregular partial observation grid $\mathcal{X}\subset\Omega$ on discrete times $t$. Please note that to simulate the challenges faced in real-world applications of the fluid environment, both in the training set and test set, the observation grid is randomly generated, and the modality $u$, $v$, and $p$ of NSE are randomly observed, where $u$ and $v$ are the velocities in $x$ and $y$ directions, respectively, and $p$ denotes the pressure.


The objective of the proposed PIR is to learn the representation $z$ of a trajectory, where the definition of trajectory is to calculate the solution of the PDE for a period of time given the PDE formula, initial conditions, and boundary conditions. Specifically, a long period of solution $\mathbf{u}([0:T], \mathbf{x})$ can be divided into many trajectories with certain initial conditions $\{\mathbf{u}([0:n],\mathbf{x}),\mathbf{u}([1:n+1],\mathbf{x}), ..., \mathbf{u}([T-n:T],\mathbf{x})\}$, where $n\in [0,T]$ is the length of the trajectory. By using the proposed PIR, a map is learned as follows:

\begin{equation}
    z(t) := f_\beta(\mathbf{u}([t:t+n],\mathbf{x})),
\end{equation}
where $\beta$ is the learnable parameters of the neural network.

\section{Methodology}  \label{method}

\subsection{Physics-informed Neural Networks} \label{sec.pinns}
We begin with a brief overview of Physics-informed Neural Networks (PINNs) \citep{karniadakis2021physics, cai2021physics} in the context of inferring the solutions of PDEs. Generally, we consider the PDE taking the form of Eqn. \ref{eqn.1}.
Following the original work of \cite{raissi2017physics}, we proceed by representing the unknown solution $\mathbf{u}(t, \mathbf{x})$ by a deep neural network with parameters $\theta$. Then, a physics-informed model can be trained by minimizing the following composite loss function:

\begin{equation}
\begin{aligned}
\mathcal{L}=\lambda_{IC} &\mathcal{L}_{IC}+\lambda_{BC} \mathcal{L}_{BC}+\lambda_{PDE} \mathcal{L}_{PDE}, \\
\mathcal{L}_{IC} = & \frac{1}{N_{IC}}\sum_{i=1}^{N_{IC}}\|\mathbf{u}\left(0, \mathbf{x}_{IC}^i\right)-g\left(\mathbf{x}_{IC}^i\right)\|^2, \\
\mathcal{L}_{BC} = & \frac{1}{N_{BC}}\sum_{i=1}^{N_{BC}}\|\mathcal{B}\left[\mathbf{u}\right]\left(t_{BC}^i, \mathbf{x}_{BC}^i\right)\|^2, \\
\mathcal{L}_{PDE} = & \frac{1}{N_{PDE}} \sum_{i=1}^{N_{PDE}}\|\frac{\partial \mathbf{u}}{\partial t}\left(t_{PDE}^i, \mathbf{x}_{PDE}^i\right) \\
& + \mathcal{N}\left[\mathbf{u}\right]\left(t_{PDE}^i, \mathbf{x}_{PDE}^i\right)\|^2,
\end{aligned}
\label{pinn loss}
\end{equation}
where $\left\{\mathbf{x}_{IC}^i\right\}_{i=1}^{N_{IC}}$, $\left\{t_{BC}^i, \mathbf{x}_{BC}^i\right\}_{i=1}^{N_{BC}}$, and $\left\{t_{PDE}^i, \mathbf{x}_{PDE}^i\right\}_{i=1}^{N_{PDE}}$ are the vertices of a fixed mesh or points that are randomly sampled at each iteration of a gradient descent algorithm. Notice that all required gradients with respect to input variables or network parameters $\theta$ can be computed via automatic differentiation \citep{griewank2008evaluating}. Moreover, the hyper-parameters $\left\{\lambda_{IC}, \lambda_{BC}, \lambda_{PDE}\right\}$ allow the flexibility of assigning a different learning rate to each individual loss term in order to balance their interplay during model training. 

\subsection{Reinforcement Learning} \label{rl}
Reinforcement Learning (RL) is a powerful approach for tackling sequential decision-making problems. In this framework, an RL agent continuously interacts with its environment, aiming to maximize the cumulative expected rewards. This problem is typically modeled as a Markov Decision Process (MDP), defined by the tuple \((S, A, P, R)\) \citep{sutton1999reinforcement}. In this context, \(S\) represents the set of states, \(A\) the set of actions, \(P\) the state transition probability matrix, and \(R\) the reward function. Within an MDP, the agent's interaction sequence forms a trajectory: \(s_1, a_1, r_1, s_2, a_2, r_2, s_3, \ldots\), where \(s_t, a_t, r_t\) denote the state, action, and reward at time step \(t\), respectively. The objective is to discover the optimal policy \(\pi^*: S \rightarrow A\) that maximizes the return. The return is typically calculated using a discount factor, as shown below:

\begin{equation}
\begin{aligned}
G_t = \sum_{t'=t}^{T} \gamma^{t'-t} r_{t'},
\label{eqn.discount}
\end{aligned}
\end{equation}
where \(T\) is the terminal time step, \(\gamma \in (0,1]\) is the discount factor, and \(r_t \in R\) is the reward at time step \(t\).

In reinforcement learning, policy gradient methods like Soft Actor-Critic (SAC) are widely used \citep{haarnoja2018soft, de2021soft}. SAC computes the value of a policy by maximizing both the expected reward and the entropy of the policy. The inclusion of entropy encourages exploration.

The objective of SAC is to maximize the entropy-regularized expected return, defined as:
\begin{equation}
\mathcal{J}(\pi) = \sum_{t=0}^{T} \mathbb{E}_{(s_t, a_t) \sim \rho_\pi} \left[ r(s_t, a_t) + \alpha \mathcal{H}(\pi(\cdot|s_t)) \right],
\end{equation}
where \(\mathcal{H}(\pi(\cdot|s_t))\) denotes the entropy of the policy, and \(\alpha\) is a hyperparameter that balances the reward and entropy.

The value functions in SAC follow the soft Bellman equations:
\begin{equation}
\begin{aligned}
Q(s_t,a_t) &= r_t + \gamma \mathbb{E}_{s_{t+1}} \big[ V(s_{t+1}) - \alpha \log \pi(a_{t+1}|s_{t+1}) \big], \\
V(s_{t+1}) &= \mathbb{E}_{a_{t+1}} \big[ Q(s_{t+1},a_{t+1}) - \alpha \log \pi(a_{t+1}|s_{t+1}) \big],
\end{aligned}
\end{equation}
where the entropy term $\alpha \log \pi$ encourages exploration.
In this work, we combine our proposed PIR with the RL algorithm to handle the control problem in complex fluid environments.

\subsection{Model Architecture}
The architecture of PIR has two components:  a physics-informed decoder and an encoder, as shown in Fig. \ref{fig.1}. 

\textbf{Physics-informed decoder:} $D_\theta(z,\mathbf{x},t)$. It is the key component of PIR, where the input is the representation $z$, coordinate $\mathbf{x}$, and time $t$, and the output is the decoded flow field $\mathbf{\hat{u}}$. The architecture applies the MLP to extract features from $z$ based on $\mathbf{x}$ and $t$, decoding them into $\mathbf{\hat{u}}$ as follows:
\begin{equation}
    \mathbf{\hat{u}}(\mathbf{x},t) = D_\theta(z,\mathbf{x},t),
\end{equation}
where $\theta$ is the learnable parameters.

\textbf{Encoder:} $E_\alpha(\mathbf{u}, \mathbf{x}, t)$. We evaluated two methods for learning the representation 
$z$: (1) direct optimization of trainable parameters and (2) an encoder-based approach where $z$ is generated by a neural network. While both methods achieved comparable performance in experiments, they differ critically in inference efficiency. The trainable parameters require re-optimizing for a few epochs when receiving new observations, introducing latency, whereas the encoder rapidly produces $z$ via a single forward pass. Given the real-time demands of the downstream control tasks, we prioritize inference speed and adopt the encoder, as it eliminates the need for repeated optimization during deployment.


\begin{figure*}[!htb]
\centering
\centerline{\includegraphics[width=1\textwidth]{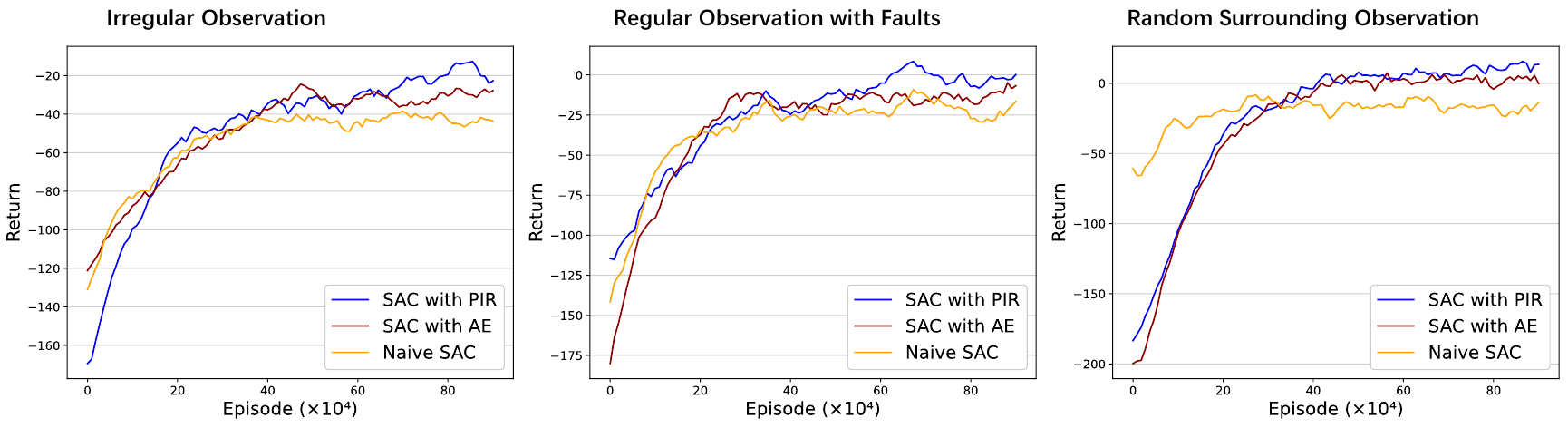}}
\caption{Return curves on the test set of the three methods during the training. We save the number of 100 checkpoints during the training and evaluate the changes of their returns. In line with the RL research community, we smooth the data when plotting.}
\label{fig.4}  
\end{figure*}

The architecture we applied to the encoder is the MLP, the same as the physics-informed decoder. Please note that our proposed PIR algorithm does not rely on a specific MLP network, and other neural network architectures, such as the Transformer, can also be used, but this is not the focus of this work. We use the same MLP to align with the structure of PINNs. The input of the encoder is $\mathbf{u},\mathbf{x},t$ and the output is $z(t)$ as follow:
\begin{equation}
    z(t) = E_\alpha(\mathbf{u}, \mathbf{x}, t),
\end{equation}
where $\alpha$ is the learnable parameters of the neural network.

\subsection{Model Learning} \label{model learning}
Based on the previous model architecture, there are two problems in jointly training the model via data loss and PDE loss. The first one is how to train the encoder to learn $z$ without the label. The second one is how to leverage the physics-informed decoder to learn the PDE dynamics. To solve the problems, we proposed a learning strategy to train two components separately instead of using the end-to-end training manner. 

Intuitively, in the case of the determined PDE dynamics, representing a trajectory requires learning initial conditions and boundary conditions. Therefore, the role of the physics-informed decoder is to learn the PDE dynamics through the PDE loss, similar to learning the neural operator solution for the PDE. Then, when the dynamics is learned, the decoder's outputs $\mathbf{\hat{u}}(\mathbf{x},t)$ can be aligned with the labels $\mathbf{u}(\mathbf{x},t)$ (given at different positions and time steps) by optimizing the encoder.
The encoder's output $z(t)$ is the representation, which is input to the physics-informed decoder along with $\mathbf{x}$ and $t$. When the $\mathbf{\hat{u}}(\mathbf{x},t)$ are consistent with the $\mathbf{u}(\mathbf{x},t)$, it is considered that a solution trajectory of PDE is uniquely determined through the $z(t)$, $\mathbf{x}$, and $t$, that is, $z(t)$ has learned the representation of this trajectory. 

We design the loss function as follows:
\begin{equation}
    \mathcal{L}(\theta,\alpha,\mathcal{D}) = \mathcal{L}_D(\theta,\alpha,\mathcal{D}) + \gamma\mathcal{L}_{PDE}(\theta, \mathcal{D}),
\end{equation}
where $\mathcal{L}_D$ is the data loss calculated on observations by MSE and $\mathcal{L}_{PDE}$ is the PDE loss calculated as Eqn. \ref{pinn loss}. $\gamma$ is the weight of the PDE loss. $\mathcal{L}_D$ is used to train the PIR end-to-end, while the gradient of $\mathcal{L}_{PDE}$ is only backpropagated to the physics-informed decoder. We also conduct experiments that use $\mathcal{L}_{PDE}$ to train the encoder. However, the performance is worse than our designed training strategy. 

\begin{table}[t]
    \centering
    \caption{Comparison of $Error_\text{consist}$ to demonstrate the consistency of $z$.}
    \label{table.1}
    \renewcommand{\arraystretch}{1.2}
    \begin{small}
    \begin{tabular}{c|cc}
    \toprule
    Method & Full modality $\downarrow$ & Missing modality $\downarrow$ \\
    \hline
     AE & 0.7625 & 0.8091 \\
    \textbf{PIR} & \textbf{0.4024} & \textbf{0.3659} \\
    \bottomrule
    \end{tabular}
    \end{small}
\end{table}

\begin{table}[t]
\centering
\caption{Comparison of $Error_\text{freq}$ to demonstrate the representation accuracy in the Fourier domain.}
\label{tab:fourier}
\renewcommand{\arraystretch}{1.2}
\begin{small}
\begin{tabular}{c|cc}
\toprule
 Method & Full modality $\downarrow$ & Missing modality $\downarrow$ \\
\midrule
 AE & 0.3187 & 0.2034 \\
\textbf{PIR} & \textbf{0.0869} & \textbf{0.0905} \\
\bottomrule
\end{tabular}
\end{small}
\end{table}

\begin{table*}[!htp]
\centering
\renewcommand{\arraystretch}{1.6}
\caption{The performance of the final RL agent in three scenarios.}
\begin{small}
\begin{tabular}{l|cc|cc|cc}
\hline
\textbf{Method} & \multicolumn{2}{c|}{\textbf{Irregular Observation}} & \multicolumn{2}{c|}{\textbf{Regular Observation with Faults}} & \multicolumn{2}{c}{\textbf{Random Surrounding Observation}} \\
\cline{2-3} \cline{4-5} \cline{6-7}
& return $\uparrow$ & success rate $\uparrow$ & return $\uparrow$ & success rate $\uparrow$ & return $\uparrow$ & success rate $\uparrow$ \\
\hline
Naive SAC & -50.0868 & 45\% & 8.1940 & 55\% & 13.3191 & 65\% \\
\hline
SAC with AE & -18.0243 & 60\% & 8.1770 & 70\% & -51.4259 & 45\%\\
\hline
\textbf{SAC with PIR} & \textbf{-11.2000} & \textbf{75\%} & \textbf{25.7971} & \textbf{80\%} & \textbf{15.9832} & \textbf{75\%} \\
\hline
\end{tabular}
\label{tab:rl_results}
\end{small}
\end{table*}

\section{Experiments} \label{exp}
\subsection{Evaluation Scenarios}
To evaluate if $z$ learned using PIR can achieve our objective and represent a fluid trajectory, we leverage two methods: direct comparison of representations' accuracy in the Fourier domain and evaluation on the control performance of multi-modal policies combined with the RL algorithm. 

\subsection{Benchmarks and Baselines} \label{benchmark}
\textbf{Benchmarks:} Before the discussion of experimental results, we first introduce the benchmarks and baselines in this subsection. 
The study delves into a real-world scenario of incompressible fluid dynamics, governed by the renowned Navier-Stokes equation (NSE), as outlined in \cite{raissi2019physics, feng2023control}. 
NSE is one of the most important PDEs that describe fluid behavior, widely applied in meteorology, oceanography, aviation, automotive, medical, and environmental fields, serving as fundamental tools for many scientific research and engineering design projects.


In our work, we consider the same dataset as \citet{raissi2019physics} and divide the long solution data into about 200 short trajectories, where the solution of NSE with $Re=100$ and the formula is described as:
\begin{equation}
\begin{aligned}
\frac{\partial u}{\partial t} + u \frac{\partial u}{\partial x} + v \frac{\partial u}{\partial y} &= -\frac{1}{\rho} \frac{\partial p}{\partial x} + \nu \left( \frac{\partial^2 u}{\partial x^2} + \frac{\partial^2 u}{\partial y^2} \right), \\
\frac{\partial v}{\partial t} + u \frac{\partial v}{\partial x} + v \frac{\partial v}{\partial y} &= -\frac{1}{\rho} \frac{\partial p}{\partial y} + \nu \left( \frac{\partial^2 v}{\partial x^2} + \frac{\partial^2 v}{\partial y^2} \right), \\
\frac{\partial u}{\partial x} + &\frac{\partial v}{\partial y} = 0,
\label{eqn.nse}
\end{aligned}
\end{equation}
$p$ is the pressure field, $u$ and $v$ are the velocity fields of $x$ and $y$ directions. $\nu$ denotes the viscosity coefficient. The last equation represents the continuity equation for the incompressible flow. In addition, the \textbf{full modality} scenario means both $u$, $v$, and $p$ can be accessed at the observation points, while the \textbf{missing modality} means that some modalities may be unobservable randomly at each point.

\begin{figure*}[!htb]
\centering
\centerline{\includegraphics[width=1.0\textwidth]{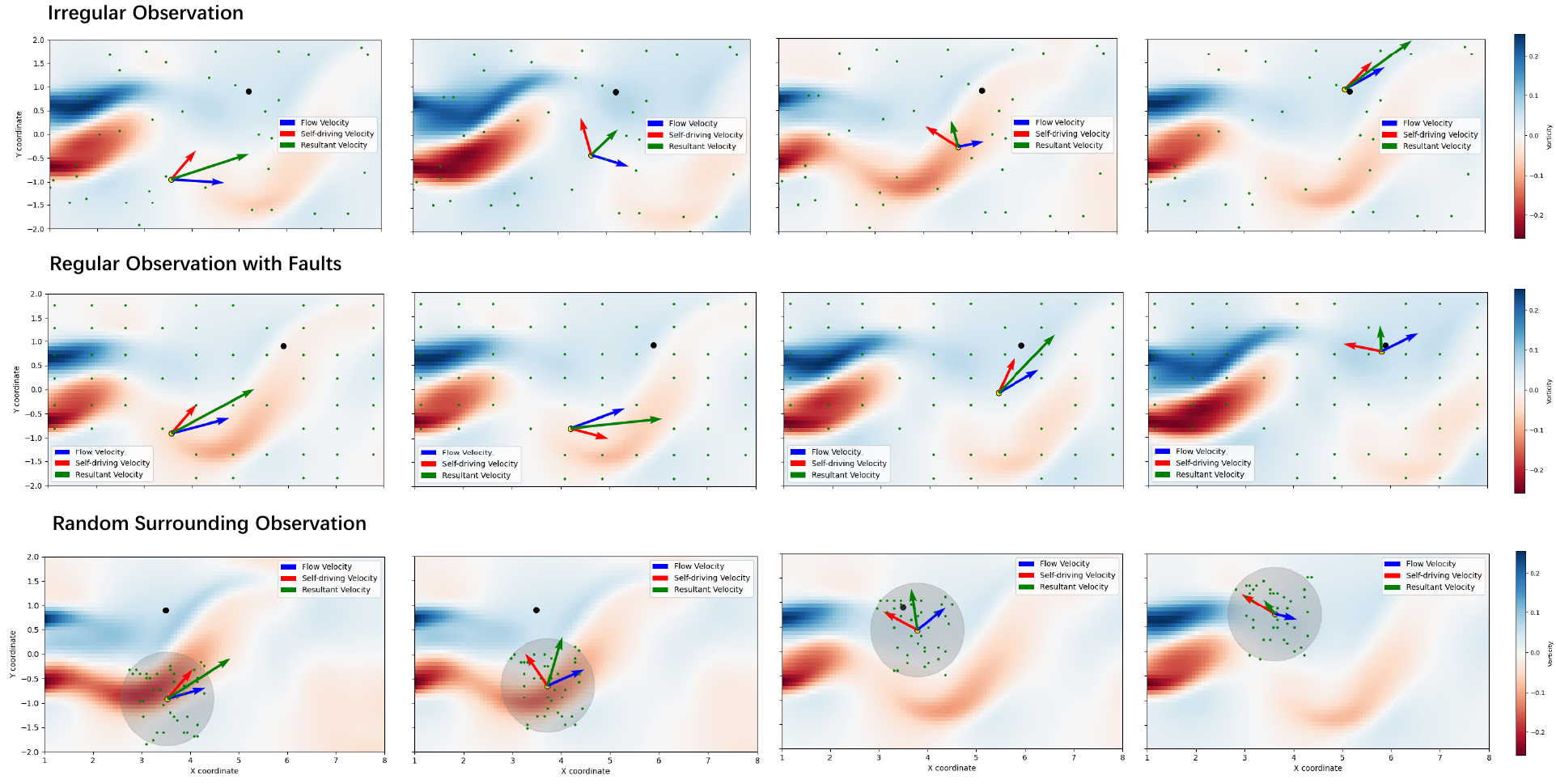}}
\caption{Visualizations of the control process in three scenarios. The red arrow is the controlled direction of self-driving and swimming velocity, the blue arrow is the flow velocity at the current position, and the green arrow is the resultant velocity.}
\label{fig.5}  
\end{figure*}

Then, we adopt the benchmark from \citet{gunnarson2021learning}, with visualizations in Fig. \ref{fig.5}.
We focus on navigating a particle robot through the unsteady von Kármán vortex street in a 2D cylinder, a scenario that poses significant challenges due to its complex, time-varying fluid dynamics. The robot must traverse from a random starting position on one side of the wake to a random target position on the opposite side, with each episode starting from a random time in the vortex shedding—testing the algorithm's adaptability to diverse conditions. Critically, the robot perceives only its relative position to the target $(\Delta x,\Delta y)$, preventing reliance on memorized flow features and necessitating real-time perception of the fluid environment. Moreover, control is limited to adjusting the robot's heading angle at a fixed self-swimming velocity ratio of $U_{swim}/U_\infty=0.8$, rendering straight paths often infeasible and forcing the robot to exploit vortex interactions for swimming (impossible to against the flow because the self-swimming velocity $U_{swim}$ cannot overcome flow velocity $U_\infty$). This setup highlights the challenge of control in unstable fluid environments.


\textbf{Baselines:} In the direct comparison of representations, the Auto-Encoder (AE) method is involved as a baseline. It is a strong baseline as it has the same architecture as PIR, but the training manner is different. AE only has data loss, while PIR also has PDE loss that is well-designed. The comparison with AE can directly illustrate our contribution. Then, in the comparison of policies with representations, a baseline is the SAC with AE that only replaces the encoder module compared with our proposed SAC with PIR. Another baseline is the naive SAC, which access the $(\Delta x,\Delta y)$ without the learned representations.

\subsection{Direct Comparison of Representations' Accuracy} \label{direct compare}
Here, we consider two aspects to show the superior representation both in full modality and missing modality scenarios. 

Firstly, we evaluate the consistency of $z$ learned by different observations from a single fluid trajectory. 
As $z$ is learned from sparse observations, representing a fluid trajectory, observations with different $\mathbf{x}$ and $t$ are expected to have consistent $z$. Thus, we compare $z$ output by the encoder with different observations. Because $z$ is in a hidden space, directly comparing is meaningless. We normalize the $z_t, t\in(1,T)$ by calculating their distance to $z_0$ ($d_t=\|z_t - z_0\|^2$). This normalization reflects the relative changes in the learned representations, such as periodicity.
Then, we calculate the relative $L_2$ error as $\text{Error}_\text{consist} = \frac{\| \mathbf{d}_t - \mathbf{d}'_t \|_2}{\| \mathbf{d}_t \|_2}$. The results are shown in Table \ref{table.1}, and we can see that $z$ learned by PIR maintains greater consistency in the representations even when there are missing modalities, with more than 47\% and 54\% reductions in full modality and missing modality, respectively. We consider this because introducing the PDE information can maintain better consistency for the model compared to using only data, and AE cannot rely on partial data features to capture global information, especially in the case of the missing modality. 

Secondly, in addition to comparing the consistency of different representations, we also compare the representations of different methods in the frequency domain. Even when mapped to the latent space, $z$ should remain consistent with the spectrum of the original data in order to reflect its modality. We use the Fourier transform on $z$ to get $\mathcal{F}(z)$ and the original flow field $\mathbf{u}$ to get $\mathcal{F}(\mathbf{u})$, and calculate the relative $L_2$ error of spectrum $Error_{freq}=\frac{\|\mathcal{F}(\mathbf{u})-\mathcal{F}(z)\|_2}{\|\mathcal{F}(\mathbf{u})\|_2}$. The results are shown in Table \ref{tab:fourier}. We can see that $z$ learned by PIR has significant improvement both in full modality and missing modality scenarios, with 72.73\% and 55.51\% improvements. This reflects that our proposed PIR better learns the features of the original flow field.




\subsection{Performance of Policies with Representations} \label{rl policy}
Beyond direct comparison of representations, their performance in control tasks is more crucial for demonstrating their capabilities. In this work, we utilize the representation $z$ as the RL state to address the multi-modal control problem. Following the benchmark introduced before, we equip the particle robot with sensors to detect velocity and pressure at sparse points, implementing three distinct configurations:

1. Irregular Observation: Sensors are positioned at fixed points within the flow field, forming an irregular mesh. The observation modalities are randomized, and each sensor location may measure only velocity, only pressure, or both velocity and pressure.

2. Regular Observation with Faults: The scenario involves sensors arranged in a regular pattern but subject to potential malfunctions during operation. To simulate real-world issues, we randomly omit observations from certain positions during evaluation. As in the first scenario, the modalities observed at each functioning sensor remain randomized.

3. Random Surrounding Observation: This challenging scenario simulates mobile sensors that fluctuate within a radius around the robot. The absolute positions of observations vary over time, adding complexity to the navigation task. As the previous configurations, the modalities observed at each position are randomized.

We apply the renowned SAC algorithm to learn the multi-modal policies. We design the reward function as follows:
\begin{equation}
    \left\{\begin{matrix} r = -\sqrt{(\Delta x+\Delta y)^2} - \omega,
    \\ r_\text{success} = 100,
    \\ r_\text{fail} = -100,
    \end{matrix}\right.
\end{equation}
where $\Delta x$ and $\Delta y$ are the relative distance from the current position to the target position. $\omega$ is the weight of the time cost. $r_\text{success}$ denotes the reward obtained when the particle robot arrives at the target successfully, and $r_\text{fail}$ means the punishment when the robot leaps out of the surrounding boundaries. The reward function means we aim to control the robot to arrive at the target from the initial position, crossing the vortex street in the shortest time step.

Figure \ref{fig.4} illustrates the return curves during evaluation across each 9,000 training episodes. The return is calculated as the cumulative sum of rewards throughout the control process, with an additional $r_\text{success}$ reward added at the final time step if the task is completed successfully, or $r_\text{fail}$ if it fails. As demonstrated in the figure, SAC with PIR achieves the highest return compared to SAC with AE and naive SAC. This indicates that the representations learned through PIR can more effectively capture the multi-modal data observed in each short trajectory, thereby enhancing the decision-making capabilities of SAC agents.
Table \ref{tab:rl_results} presents the performance of the last agents for each method. The results demonstrate that our proposed approach achieves improvements of 20.0\%, 214.8\%, and 20.0\%, respectively, compared to the best baseline. It demonstrates a significant improvement in the scenario of multi-modal observations when sensors have random faults. There are also success rates of each method during the evaluation in Table \ref{tab:rl_results}, which shows our proposed SAC with PIR always has the highest success rate, regardless of the scenario.
Furthermore, Figure \ref{fig.5} provides visualizations of the control process.

\section{Limitations and Future Works}
This work proposes a novel idea of representing a trajectory by encoding observations with PDE information. But this work retains many aspects that can be expanded in the future. Firstly, due to the lack of benchmarks of real-world measured data or real-world experimental platforms, we only use numerical data to simulate the challenges encountered in practice. Secondly, the neural network structure in PIR can be replaced with Transformer, GNNs, etc., as this algorithm can be applied to any neural network.

\section{Conclusion}
In conclusion, our proposed PIR addresses the challenge of leveraging random and sparse observations for control in complex fluid environments. PIR uses PDE information and sparse data to represent the fluid dynamics. Our experiments show that PIR maintains superior consistency with ground truth features compared to baseline methods, even in the presence of missing modalities. Moreover, PIR has been proven successful in the multi-modal control, enabling particle robots to utilize the learned representation more quickly and accurately, navigating complex environments like vortex streets from random starting points to random targets. It highlights the potential of PIR in enhancing control policies for fluid systems with random observations.

\bibliography{pir}

\end{document}